\pdfoutput=1
\documentclass[letterpaper]{article} %
\usepackage[preprint]{aaai2027}  %
\usepackage[hyphens]{url}  %
\usepackage{graphicx} %
\usepackage{natbib}  %
\usepackage{caption} %
\usepackage{amsmath}
\usepackage{amssymb}
\usepackage{booktabs}
\usepackage{multirow}
\usepackage[table]{xcolor}
\definecolor{atfblue}{RGB}{228,241,254}%
\definecolor{atfqwenA}{RGB}{246,235,250}%
\definecolor{atfqwenB}{RGB}{233,235,252}%
\definecolor{atfgraylt}{RGB}{235,235,235}%
\definecolor{atfpink}{RGB}{255,226,238}%
\definecolor{atforange}{RGB}{255,238,214}%

\title{ATFlash: Per-RoPE-Wavelength Attention Windows for Compute/Memory-Efficient LLM Inference}

\author{
    Shun-ichiro Hayashi\textsuperscript{\rm 1},
    Daichi Mukunoki\textsuperscript{\rm 2},
    Tetsuya Hoshino\textsuperscript{\rm 2},
    Takahiro Katagiri\textsuperscript{\rm 2}
}
\affiliations{
    \textsuperscript{\rm 1}Graduate School of Informatics, Nagoya University\\
    \textsuperscript{\rm 2}Information Technology Center, Nagoya University\\
    hayashi@hpc.itc.nagoya-u.ac.jp, \{mukunoki, hoshino, katagiri\}@cc.nagoya-u.ac.jp
}

\begin{document}

\maketitle

\begin{abstract}
The attention score with rotary position embeddings (RoPE) decomposes exactly into a sum over its 2D-rotation frequency pairs, and each pair's wavelength limits how far it can discriminate position.
Aligned with this structure, we propose the per-RoPE-wavelength distance window: it prunes the query--key inner-product terms beyond a wavelength-proportional distance.
Unlike a sliding window, every key remains reachable, at least through the low-frequency pairs. The reduction rate is input-independent, with a closed form logarithmic in the sequence length $N$, in contrast to dynamic-sparse methods like MInference.
Such token-level selection is orthogonal to our frequency-level pruning. The window can therefore be applied on top of those methods.
On Qwen2.5-0.5B and Llama-3.2-3B, the window prunes 37--48\% of the query--key inner-product terms within each model's native context length.
Relative to full attention, the top-1 match rate stays at 96--98\% and the mean output-distribution KL at the $10^{-3}$-nat level on LongBench-v2 contexts.
We examine absolute scores on long-context benchmarks such as RULER, OpenAI-MRCR, LongCodeQA, and $\infty$Bench: they are broadly preserved.
We implement the window as a slice of the query--key contraction axis, leaving the online-softmax recurrences untouched, and port it with minimal diffs into the released FlashAttention-4 prefill and FlashInfer decode.
On RTX PRO 6000 with Llama, both ports outpace stock with gains growing with context length, up to $1.29\times$ at 128K.
End to end on Qwen2.5-7B-1M, with 57\% of the inner-product terms pruned, the speedup reaches $1.31\times$ at a 1M-token context.
\end{abstract}

\section{Introduction}
\label{sec:intro}

Inference for large language models (LLMs) is increasingly limited by the compute of attention and by the memory bandwidth of the KV cache---the intermediate state kept for past tokens---as contexts grow longer and multi-agent deployments spread.
FlashAttention~\citep{fa2} (hereafter FA)-class kernels are the fastest implementations that carry out this computation \textbf{exactly}, but being exact, they do not reduce the amount of computation itself.
Many approximation methods that cut computation suffer from one of the following: (a)~the amount of reduction is input-dependent and cannot be predicted in advance; (b)~they discard tokens or KV, so the dropped information becomes structurally unreachable; or (c)~they are tightly coupled to a particular kernel implementation and cannot be carried to other implementations.

\begin{figure}[t]
\centering
\includegraphics[width=0.8\columnwidth]{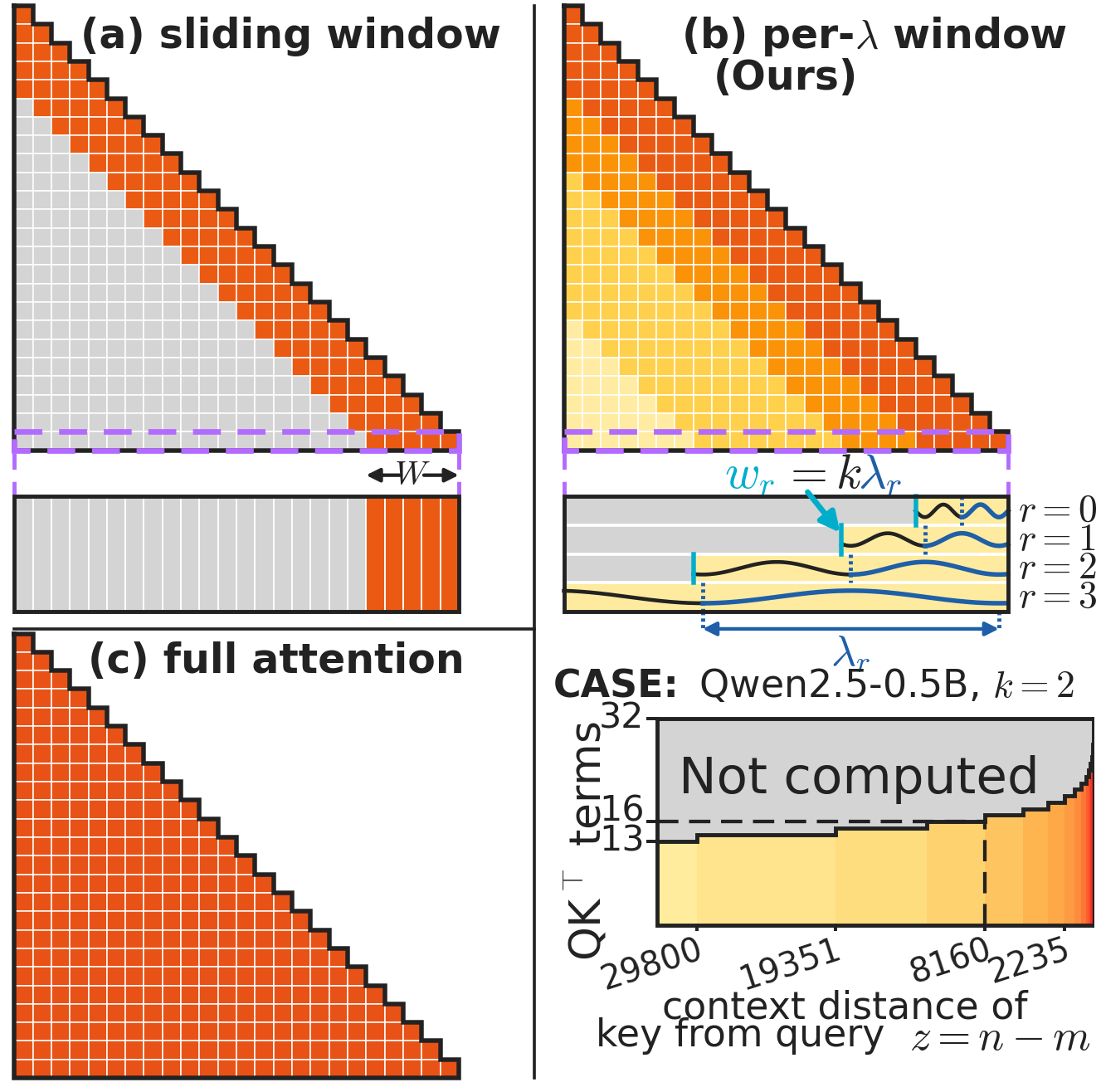}%
\caption{Contrast of a sliding window, the per-RoPE-wavelength distance window, and full attention (conceptual diagram, causal attention). (a): out-of-window tokens disappear (gray). (b): the same row is split vertically by frequency pair $r$ and truncated by the wavelength-proportional window. CASE: actual count of inner-product terms for the last query row (Qwen2.5-0.5B, $k{=}2$, $N{=}32768$).}%
\label{fig:swavspf}
\end{figure}

Our starting point is that the attention score with rotary position embedding (RoPE)~\citep{rope} decomposes \textbf{exactly} into a sum of per-frequency-pair contributions (Section~\ref{sec:method}).
Each pair has its own wavelength $\lambda_r$; high-frequency pairs complete one period in a few tokens, and low-frequency pairs in millions of tokens.
How far each component can carry positional information is determined by model constants (the RoPE base and dimension), independent of the input.
Aligned with this structure, imposing a wavelength-proportional distance window $w_r=\min(k\lambda_r,N)$ on each frequency pair to prune the terms of the score sum is the \textbf{per-RoPE-wavelength distance window}---the core of our proposed method, ATFlash (Figure~\ref{fig:swavspf}).
Unlike conventional windows that drop tokens, every token continues to be referenced through the low-frequency pairs whenever $N<k\lambda_{\max}$.

We make four contributions.
\begin{itemize}
\item \textbf{Method}: We propose a per-RoPE-wavelength distance window aligned with the exact pair decomposition of the RoPE score. The window can be implemented as a slice of the reduction axis of online softmax, and because it discards no token it can be inserted while preserving the skeleton and numerical path of FA-class kernels. It has a single free parameter, the number of retained periods $k$, and requires neither training nor hand-tuning (Section~\ref{sec:method}).
\item \textbf{Theory}: We derive the reduction rate in closed form. The reduction is logarithmic in $N$, and for $N\ge k\lambda_{\max}$ it asymptotes to $O(1)$ terms per query and $O(N)$ overall. Measured reduction matches the closed form at every point (within the 1--2-point integral-approximation error), so reduction is predictable before running (Section~\ref{sec:theory}).
\item \textbf{Implementation and portability}: We port the window with minimal diffs into the released FlashAttention-4 (hereafter FA4) prefill and the FlashInfer decode, and on RTX PRO 6000 (sm\_120, GDDR7) each surpasses its stock implementation at all evaluated Llama context lengths while remaining bit-identical with the window off. Recovery of the theoretical ceiling on Llama lines up at 85--96\% across the two ports, confirming that the loss in converting theoretical reduction into real time is not implementation-specific (Sections~\ref{sec:impl}--\ref{sec:speed}).
\item \textbf{Two-layer robustness verification}: We show that the method maintains 37--48\% reduction in the non-extrapolation regime, a Kullback--Leibler (KL) divergence on the order of $10^{-3}$, and 100\% recall on in-text needle retrieval, with a two-order-of-magnitude gap over a sliding window at equal compute, at two layers---distribution and capability (Sections~\ref{sec:eval-quality}--\ref{sec:eval-ability}). Benchmark scores are broadly preserved; the exception is retrieval over structureless strings, where $k$ acts as a dial---at uniform $k{=}64$, still pruning 29\% of the terms, the worst case returns to the full-attention score.
\end{itemize}

\section{Related Work}
\label{sec:related}

\textbf{Exact FA-class kernels.}
FA computes exact attention without materializing the score matrix, using tile-wise online softmax~\citep{onlinesoftmax} (sequential updates of a running max and a normalizer); from FlashAttention-2~\citep{fa2} onward the family has specialized by hardware generation (Hopper in FA3~\citep{fa3}, Blackwell data-center in FA4~\citep{fa4}).
This family deepens hardware fit while retaining all inner-product terms, and is orthogonal to our method, which selectively drops terms---we demonstrate in Section~\ref{sec:impl-port} that the two can be combined.

\textbf{The frequency structure of RoPE.}
RoPE~\citep{rope} applies a rotation with a geometrically spaced angular velocity $\theta_r$ for each dimension pair.
The structure of the wavelength $\lambda_r=2\pi/\theta_r$---high frequency for short distance, low frequency for long distance---and the division of roles---high frequency for position, low frequency for meaning---are widely observed~\citep{barbero}.
On the other hand, it is also known that when $q,k$ are isotropic the expectation of the RoPE contribution does not depend on distance~\citep{barbero}, so one cannot justify a window by assuming a \textbf{distance decay} of the score.
What our method relies on is not decay but the \textbf{wavelength} (Section~\ref{sec:pair}), which is fundamentally different from conventional windows based on a decay assumption.

\textbf{Sparse and windowed attention.}
The sliding window~\citep{longformer}, attention sinks with a recent window~\citep{streamingllm}, and KV eviction by cumulative scores~\citep{h2o} reduce computation, but all impose the same window or selection on every frequency, so the dropped tokens become structurally unreachable.

\textbf{Input-dependent dynamic sparsification.}
A training-free line is active here: MInference~\citep{minference} is the standard baseline with offline search over vertical-slash patterns, FlexPrefill~\citep{flexprefill} and XAttention~\citep{xattn} make the selection online and lighter, and SpargeAttention~\citep{sparge} combines block sparsity with quantization.
Approaches that use the RoPE frequency structure as a cue have also appeared: FASA~\citep{fasa} predicts token importance from dominant frequency components and dynamically selects KV entries at decode, and TriAttention~\citep{triattn} compresses KV with a trigonometric distance preference; both use frequencies only as a \textbf{selector}, and the surviving tokens are computed at all frequencies.
In all of these the selection is input-dependent, so the reduction is not fixed before execution.

\textbf{Training-based methods.}
DSA~\citep{dsa} builds a learned selector into pretraining, and SFA~\citep{sfa} prunes inner-product terms with learned codes.
Their units of reduction differ, but both presuppose training the model itself and cannot be retrofitted.

Our method carries none of the constraints (a)--(c) of Section~\ref{sec:intro}: the window is determined in closed form from the wavelengths and hence input-independent, and it is training-free and retrofittable onto the standard attention layers of any pretrained RoPE model.
Like SFA, our method prunes inner-product terms, but it derives the window analytically from the wavelengths and learns no codes.

\textbf{Customizable attention frameworks.}
The customization axes of FlexAttention~\citep{flexattn} and FlashInfer~\citep{flashinfer} run along the token (row/column) direction of the score matrix, whereas our pruning operates \textbf{inside} each inner product---along the frequency-pair axis of the head dimension---which neither programming model can currently express.
We realize it by minimal-diff insertion into the released FA4 and FlashInfer.

\section{Method: Per-RoPE-Wavelength Windows}
\label{sec:method}

\subsection{Pair Decomposition of the RoPE Score}%
\begin{figure}[t]
\centering
\includegraphics[width=0.9\columnwidth]{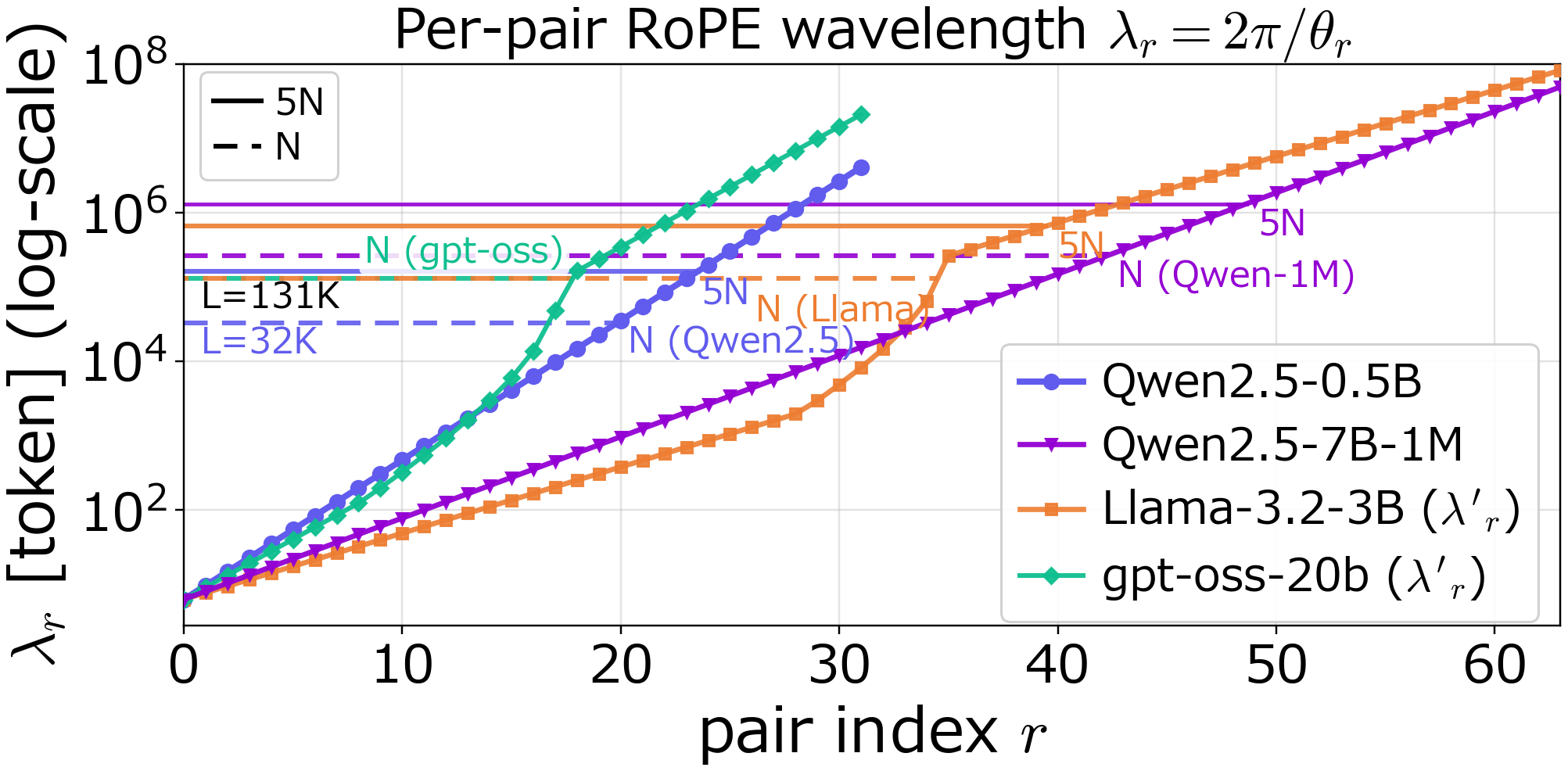}%
\caption{The wavelength $\lambda_r=2\pi/\theta_r$ of each pair (log axis). Qwen2.5-0.5B, Qwen2.5-7B-1M, Llama-3.2-3B (effective $\lambda'_r$ after llama3-style scaling), and gpt-oss-20b (effective $\lambda'_r$ under YaRN $\times$32) are shown together; horizontal lines mark each model's training length (dashed) and the $5\times$ extrapolation guide (solid).}
\label{fig:lambda}
\end{figure}

\label{sec:pair}
For a query $q$ (at position $n$) and a key $k$ (at position $m$) with head dimension $d$, RoPE applies a two-dimensional rotation with angular velocity $\theta_r=B^{-2r/d}$ ($B$ is the base) to each dimension pair $r\in\{0,\dots,d/2-1\}$.
Writing the rotated query and key as $\tilde q,\tilde k$, the score depends only on the relative position $z=n-m$ and can be written \textbf{exactly} as a sum of per-pair contributions,
\begin{equation}
\frac{\tilde q^\top \tilde k}{\sqrt d}=\!\frac{1}{\sqrt{d}}\!\sum_{r}\Delta x_{m,r},\;
\Delta x_{m,r}=a_r\cos(z\theta_r)+b_r\sin(z\theta_r)
\label{eq:pair}
\end{equation}
where $a_r,b_r$ are coefficients determined by the components of $q$ and $k$.
The \textbf{wavelength} of each pair,
\begin{equation}
\lambda_r = 2\pi/\theta_r
\label{eq:lambda}
\end{equation}
is the number of tokens over which $\cos,\sin$ complete one period along the distance $z$, and beyond it a single pair can no longer discriminate position (aliasing).
That is, high-frequency (small $\lambda_r$) pairs carry positional information only nearby, while low-frequency (large $\lambda_r$) pairs remain informative out to long distances (Figure~\ref{fig:lambda}).
The truncation of our method therefore assumes no score decay; it aligns with the upper limit of the distance over which each pair can discriminate position.
We cast this as a hypothesis, that a pair's term becomes redundant at distances where the pair cannot discriminate position, and verify it on output distributions and capabilities from Section~\ref{sec:eval-quality} onward.

\subsection{Window Rule}
\label{sec:window}
We therefore vary the distance window per frequency:
\begin{equation}
w_r = \min(k\,\lambda_r,\ N)
\label{eq:window}
\end{equation}
and at cell $(n,m)$ we keep in the score sum~(\ref{eq:pair}) only the pairs satisfying $z\le w_r$.
The number of retained periods $k$ is the \textbf{only free parameter} of this rule, the wavelengths $\lambda_r$ being model constants and $N$ the input length; because the dependence on $k$ is only logarithmic (Section~\ref{sec:theory}), we default to $k{=}2$.
The parameter thus acts as a dial between attention accuracy and compute that can be set at deployment time, much as weight precision is chosen per deployment, and its cost is known in closed form before running.
High frequencies get short windows and low frequencies get long windows, producing a staircase-like profile in which farther cells have fewer active pairs (fewer inner-product terms)---for Qwen2.5-0.5B at $N{=}32768$, $k{=}2$, from 32 terms at $z{=}0$ down to 13 at the farthest distance shown (Figure~\ref{fig:swavspf}, CASE).
Crucially, distant tokens continue to be referenced through the low-frequency pairs; our method \textbf{drops terms of the score sum per frequency rather than dropping tokens}.
Full token retention holds in the range $N<k\lambda_{\max}$, where the lowest-frequency window covers the sequence.
A sliding window corresponds to the special case of imposing the highest-frequency window on all pairs (Figure~\ref{fig:swavspf}(a)(b)).

The window table (the list of window widths for each pair) is generated from the effective frequencies in the model config.
Frequency transforms such as YaRN or llama3-style rope\_scaling are reflected in the window rule through the effective $\theta'_r$ (and hence $\lambda'_r$), with no per-scaling-scheme hand-tuning (Figure~\ref{fig:lambda}).

\subsection{Implementation as a Slice of the Reduction Axis of Online Softmax}
\label{sec:slice}
Because the window~(\ref{eq:window}) does not depend on the input and is fixed in advance, the set of active pairs per tile is determined statically.
Out-of-window pruning is implemented not as a multiplicative mask but as a shortening (\textbf{slice}) of the reduction length of the inner product.
As the distance $z$ grows, pairs fall out of the window from the high-frequency side, so the active-pair set is a contiguous slice on the low-frequency side; we round its boundary to a multiple of the K-direction atom (16 elements) of the matrix multiply-accumulate (MMA) instruction.
As a result, the tiling, running max, and normalizer of the FA-class kernel are left \textbf{entirely unchanged}, and only the effective inner-product length of each tile shrinks statically (tokens and $V$ are loaded in full; nothing is discarded).
The low-frequency terms that remain even at the farthest distance fit within a single MMA tile (Figure~\ref{fig:swavspf}, CASE), preserving Tensor Core fill density.

\begin{figure*}[t]
\centering
\includegraphics[width=1.0\textwidth]{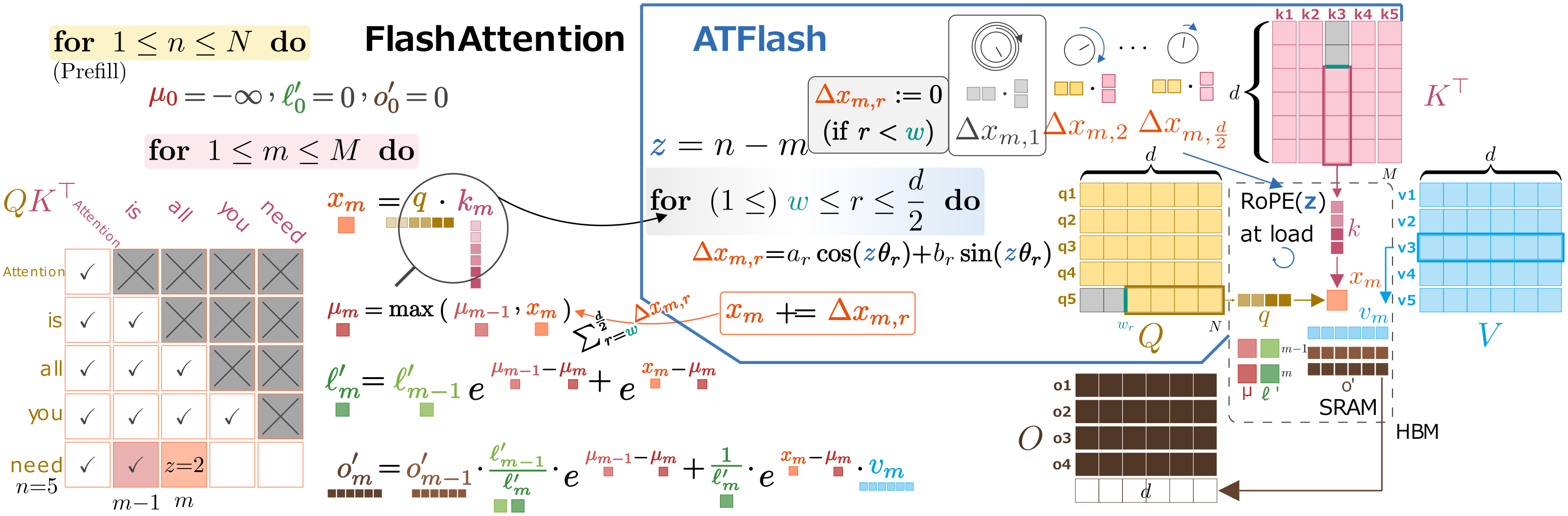}%
\caption{FA's online-softmax loop and where the per-RoPE-wavelength distance window is inserted (a full view including the double loop of prefill; decode corresponds to the last query row $n{=}N$ only). Left: FA's recurrences (running max $\mu$, normalizer $\ell'$, output $o'$) and the causal mask. Center: the pair decomposition of the score $x_m$ and the window predicate (gray box). Right: the memory-side behavior (gray $=$ out-of-window components, not loaded for this reduction).}
\label{fig:fa}
\end{figure*}

As in Figure~\ref{fig:fa}, the window insertion merely advances the start of the pair loop to the window boundary.
More important is the memory side (right of the figure): out-of-window pair components not only skip computation but are \textbf{not loaded into SRAM at all} for the relevant parts of $K^\top$ and $q$ (gray in the figure).
Thus the window cuts the QK compute and the transfer of $K^\top$ and $q$ at the same ratio (the transfer of $V$ is not reduced).
This ``slice of the reduction axis'' form underlies the cross-implementation portability of Section~\ref{sec:impl}.

\section{Theory: Closed-Form QK Reduction Rate}
\label{sec:theory}

Because the window is fixed in advance, the amount of reduction is deterministic and can be analyzed without running.
We measure the computational cost by the \textbf{number of terms of the query--key inner product} in score computation: whereas full attention requires $d/2$ pairs of multiply-accumulate per causal cell, with the window only the pairs with $z\le w_r$ are computed.
We define the reduction rate as ``$1-$(number of terms retained inside the window)$/$(number of causal cells $\times\,d/2$)''.
This definition is a term-count proxy that excludes softmax, the $V$ product, and memory transfer; wall-clock time is measured in Section~\ref{sec:speed}.

Approximating the pair index by a continuous variable $\rho$, the window width becomes the geometric progression $w(\rho)=w_{\min}B^{2\rho/d}$ ($w_{\min}=2\pi k$), and the number of retained terms over the interval $[n_0,n_1]$ can be written as the double integral
\begin{equation}
S(n_0,n_1)=\int_{n_0}^{n_1}\!\!\int_{0}^{d/2}\min\bigl(w(\rho),n\bigr)\,d\rho\,dn
\label{eq:S}
\end{equation}
which solves in closed form by splitting the integral at the saturation boundary $\rho^{*}(n)=\tfrac{d}{2}\,\ln(n/w_{\min})/\ln B$ (where $w(\rho^{*}){=}n$).
For the whole prefill ($[0,N]$, $N\gg w_{\min}$),
\begin{equation}
\mathrm{reduction}_{\mathrm{prefill}}\approx\frac{\ln(N/w_{\min})-3/2}{\ln B}
\label{eq:redprefill}
\end{equation}
For Qwen2.5 at $N{=}32$K the substituted value is 46.1\% and the exact discrete sum is 47.6\%; the difference is the integral-approximation error.
The reduction rate for decode (the last row) has the same form, and its difference from the whole prefill is only the constant $1/(2\ln B)$ (3.6 points for $B{=}10^6$).

\begin{figure}[t]
\centering
\includegraphics[width=1.0\columnwidth]{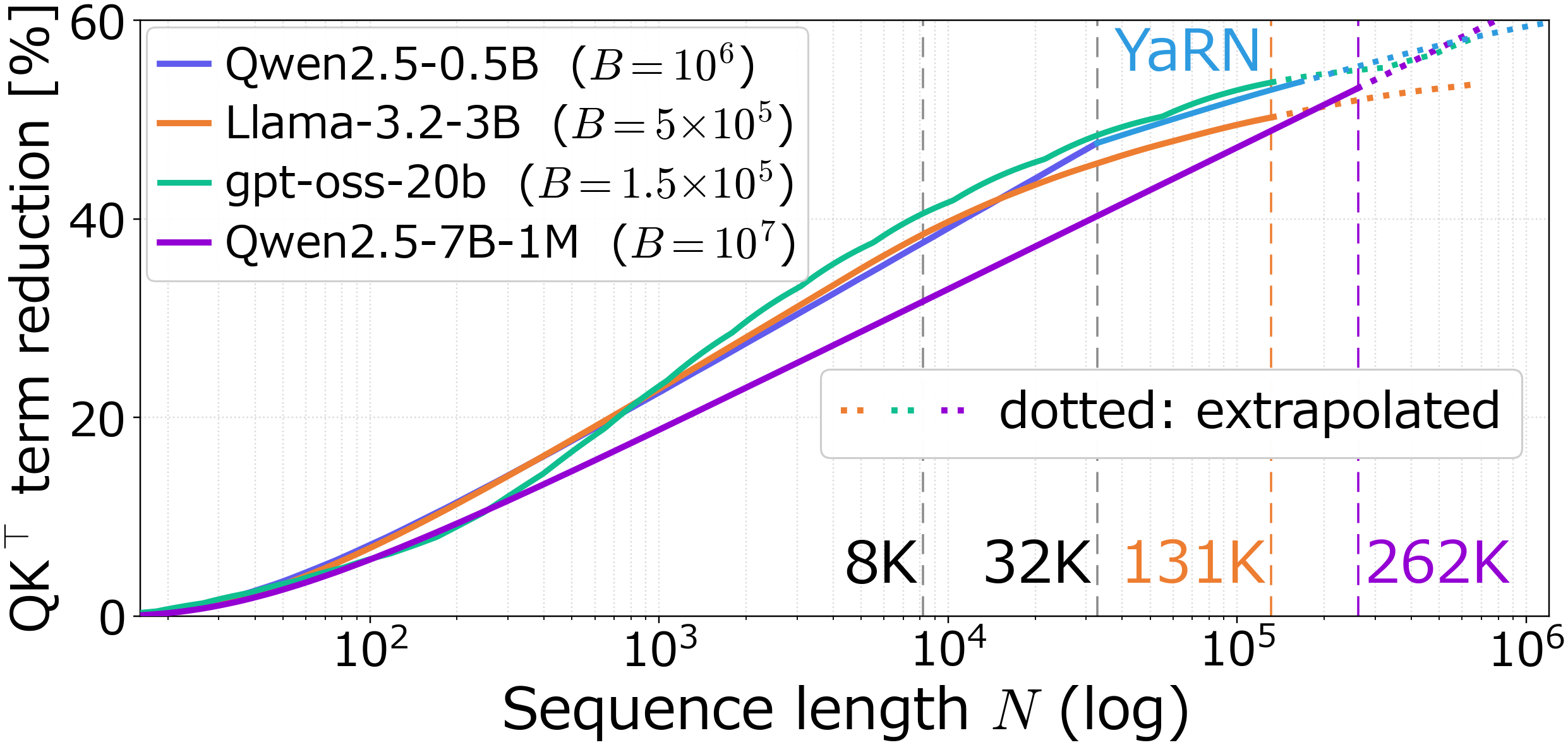}%
\caption{QK term reduction against sequence length $N$: the closed form of Eq.~(\ref{eq:redprefill}); agreement with measurement is in Table~\ref{tab:longbench}. Light blue: the closed form for Qwen2.5 ($B{=}10^6$) under YaRN at the minimal scale per $N$.}
\label{fig:saving}
\end{figure}

Three structural facts can be read from this closed form.
(i)~The reduction rate is \textbf{logarithmic} in $N$ (slope $1/\ln B$) and deepens for longer contexts.
(ii)~The dependence on $k$ is also logarithmic; doubling $k$ costs only $\ln 2/\ln B$ of the reduction---so headroom on the quality side can be bought cheaply.
(iii)~For $N\ge k\lambda_{\max}$ the $\min$ of every window settles on the $k\lambda_r$ side (no window is clipped by $N$), the number of retained terms reaches the constant $\sum_r w_r$ independent of $N$, and it becomes \textbf{$O(1)$ terms per query and $O(N)$ over the whole sequence} (Qwen2.5, $k{=}2$: 64\% at $N{=}200$K, 76\% reduction at 1M).
Note that $k\lambda_{\max}$ is orders of magnitude beyond practical context lengths (Figure~\ref{fig:lambda}); all experiments in this paper lie in the full-retention regime $N<k\lambda_{\max}$---the $O(N)$ behavior is an asymptotic guarantee.
Figure~\ref{fig:saving} shows the closed-form curves; Llama-3.2-3B is counted on its effective wavelengths after llama3-style scaling, and gpt-oss-20b on its full-attention layers only (a hybrid architecture interleaving sliding-window and full-attention layers). The QK-pruned column of Table~\ref{tab:longbench} (measured counts on real $q,k$) matches the closed form within the integral-approximation error at every point.

\section{Implementation: Window Ports}%
\label{sec:impl}
\label{sec:impl-port}

We implement the method as minimal-diff ports into the released implementations: FA4 for prefill and FlashInfer for decode.
The FA4 port changes 4 files ($+296/-13$ lines) in the forward path of the CuTeDSL implementation.
The base is the \textbf{released upstream main} (public commit \texttt{5835c73}).
The insertion truncates the QK reduction of the unmasked band to the leading MMA-K chunks of the effective inner-product length per distance band and narrows the asynchronous load of K columns to the same width (a prefix-only load for the contiguous slice on the low-frequency side); it does not touch the masked band, softmax, the PV product, or the online accumulation.
Because all added branches resolve to compile-time constants, the window-off path generates the same trace as upstream---a \textbf{bit-identical design}.

Because FA4's decode path drops occupancy at query length 1, the FA4 port targets prefill.
For decode we ported the same window into the decode kernel of FlashInfer 0.6.13 (the released wheel).
This kernel partitions the KV sequence across blocks and merges the partial online-softmax states afterward; the loop inside each split is identical to Figure~\ref{fig:fa}.
The only changes are adding the window condition to the asynchronous K-load predicate and nullifying the partial QK dot under the same predicate (about 100 lines in effect); the window-off specialization follows the stock code path.
The FlashInfer port likewise resolves at compile time, and we confirm in measurement that both ports are bit-identical to stock with the window off.
\section{Evaluation of Output Quality}
\label{sec:eval}

This section verifies output quality at 2 layers: a relative comparison against full attention (preservation rates; Section~\ref{sec:eval-quality}), and the absolute scores of the benchmarks themselves (Section~\ref{sec:eval-ability}).
The experiments run on NVIDIA GH200 and H100 GPUs.

\subsection{Fidelity: Preserved Output Distributions}
\label{sec:eval-quality}
\textbf{Setup.} We embedded the window into PyTorch's eager attention implementation (transformers 4.57.1; 5.5.3 for RULER and gpt-oss-20b) and compared it against full attention (window off) on real models and real tasks, in fp32 (bf16 for gpt-oss-20b).
The main fidelity metric is the average KL divergence (nats) from the full-attention model's output distribution, alongside the argmax match rate (top-1).

\textbf{Results.} Table~\ref{tab:longbench} shows sequence-length scaling in the LongBench-v2 context.
The wavelength-proportional $k{=}2$ window reduces the number of inner-product terms by \textbf{37--48\%} in the \textbf{non-extrapolation regime} (within each model's native context; up to 32K for Qwen2.5-0.5B and up to 64K for Llama) while keeping top-1 at 96--98\% and KL on the order of $10^{-3}$\,nats.
In the YaRN extrapolation regime the reduction grows to 53.8\% (Qwen 160K) and top-1 declines gently to 88.1\%.
This decline mainly reflects the full-attention model itself beginning to break down in the extrapolation regime, not a long-context collapse of the window's pruning---within the verification region, top-1 and KL stay nearly flat even as the reduction share grows (Table~\ref{tab:longbench}).
\begin{table}[tb]
\centering
{\footnotesize%
\setlength{\tabcolsep}{2.5pt}
\begin{tabular}{lrcrrr>{\columncolor{atfblue}}r}%
\toprule
LongBench-v2 & $N$ & QK-pruned & top-1 & KL div & full & ours \\%
\midrule
\multirow{5}{*}{Qwen2.5-0.5B}
   & 8K   & 37.6\% & 97.4\% & 0.0020 & 0.36 & \textbf{0.38} \\
 & 16K  & 42.6\% & 96.9\% & 0.0029 & 0.26 & \textbf{0.30} \\
 & 32K  & 47.6\% & 96.2\% & 0.0045 & 0.26 & \textbf{0.34} \\
\cmidrule(lr){2-7}
 & 64K$^{\dagger}$  & 50.4\% & 94.9\% & 0.0174 & 0.16 & \textbf{0.24} \\
 & 160K$^{\ddagger}$ & 53.8\% & 88.1\% & 0.0954 & 0.08 & \textbf{0.18} \\
\midrule
\multirow{4}{*}{Llama-3.2-3B}
   & 8K  & 38.5\% & 97.8\% & 0.0026 & \textbf{0.28} & 0.22 \\
 & 16K & 42.5\% & 96.6\% & 0.0055 & 0.34 & 0.34 \\
 & 32K & 45.6\% & 96.4\% & 0.0056 & 0.24 & 0.24 \\
 & 64K & 48.1\% & 97.0\% & 0.0055 & 0.16 & \textbf{0.18} \\
\bottomrule
\end{tabular}}
\caption{Sequence-length scaling ($k{=}2$, all query positions). QK-pruned $=$ share of inner-product terms pruned. full/ours $=$ accuracy on official LongBench-v2 4-choice questions, 50 per band, scored by choice-letter likelihood with contexts tail-truncated to each $N$; not comparable to the official generation-based scoring. $^{\dagger}$YaRN $s{=}2$, $^{\ddagger}$YaRN $s{=}5$.}
\label{tab:longbench}
\end{table}

\begin{table*}[t]
\centering
{\footnotesize
\setlength{\tabcolsep}{3pt}
\begin{tabular}{lcccccccccccc}
\toprule
model & \multicolumn{8}{c}{Qwen2.5-7B-1M ($d{=}128$)} & \multicolumn{3}{c}{Qwen2.5-0.5B ($d{=}64$)} & gpt-oss-20b \\
\cmidrule(lr){2-9}\cmidrule(lr){10-12}\cmidrule(lr){13-13}
benchmark & MRCR & LongCode & \multicolumn{6}{c}{$\infty$Bench} & \multicolumn{3}{c}{RULER} & GSM8K \\
\cmidrule(lr){4-9}\cmidrule(lr){10-12}
subset & custom & QA & 10 task & \multicolumn{2}{c}{Math.Find} & \multicolumn{3}{c}{Retr.KV$^{*}$} & 4 task & \multicolumn{2}{c}{MK-NIAH} & 0-shot \\%
\cmidrule(lr){5-6}\cmidrule(lr){7-9}\cmidrule(lr){11-12}
context & 66--160K & 31--245K & $\le$262K & \multicolumn{2}{c}{$\sim$117K} & \multicolumn{3}{c}{$\sim$169K} & 16K & \multicolumn{2}{c}{16K} & $\sim$1.2K \\%
$n$ & 165 & 233 & 1425 & 350 & 350 & \multicolumn{3}{c}{20} & 1200 & 300 & 300 & 500 \\%
\cmidrule(lr){1-13}
$k$ & 2 & 2 & 2 & \cellcolor{atfgraylt}2 & \cellcolor{atforange}16$\to$2 & \cellcolor{atfgraylt}2 & \cellcolor{atforange}16$\to$2 & \cellcolor{atfpink}64 & 2 & \cellcolor{atfgraylt}2 & \cellcolor{atforange}16$\to$2 & 2 \\
QK-pruned (ours) & 45.8\% & 46.3\% & 48.6\% & \cellcolor{atfgraylt}48.2\% & \cellcolor{atforange}37.8\% & \cellcolor{atfgraylt}50.5\% & \cellcolor{atforange}40.3\% & \cellcolor{atfpink}29.0\% & 42.6\% & \cellcolor{atfgraylt}42.6\% & \cellcolor{atforange}29.8\% & 9.8\% \\
\midrule
full & \textbf{0.204} & 0.622 & \textbf{0.393} & \multicolumn{2}{c}{0.334} & \multicolumn{3}{c}{\textbf{0.750}} & 20.6 & \multicolumn{2}{c}{\textbf{96.7}} & 0.908 \\%
\rowcolor{atfblue}
ours & 0.181 & \textbf{0.627} & 0.386 & \cellcolor{atfgraylt}0.334 & \cellcolor{atforange}\textbf{0.343} &\cellcolor{atfgraylt}0.300 & \cellcolor{atforange}0.350 & \cellcolor{atfpink}\textbf{0.750} & \textbf{21.3} & \cellcolor{atfgraylt}93.7 & \cellcolor{atforange}96.0 & \textbf{0.920} \\%
\midrule
MInf & \textbf{0.229} & \textbf{0.670} & \textbf{0.389} & \multicolumn{2}{c}{\textbf{0.349}} & \multicolumn{3}{c}{\textbf{0.800}} & --- & --- & --- & --- \\
\rowcolor{atfblue}
ours$+$MInf & 0.218 & 0.640 & \cellcolor{white}--- & \cellcolor{white}--- & \cellcolor{white}--- & \cellcolor{atfgraylt}0.350 & \cellcolor{white}--- & \cellcolor{atfpink}\textbf{0.800} & \cellcolor{white}--- & \cellcolor{white}--- & \cellcolor{white}--- & \cellcolor{white}--- \\
\bottomrule
\end{tabular}
}
\caption{Absolute benchmark scores. Bold $=$ best within each rule-separated block; the gray/orange cell pairs contrast the fixed $k{=}2$ window with the variable-$k$ profile; $^{*}$ $=$ significant drop for the window. Conventions and symbols: Section~\ref{sec:eval-ability}.}
\label{tab:absbench}
\end{table*}

\subsection{Capability: Absolute Benchmark Scores}
\label{sec:eval-ability}
Preserved distributions would mean little if capabilities broke.
This section scores the window on benchmarks in absolute terms: if the terms beyond a pair's wavelength carry noise rather than positional signal (the hypothesis of Section~\ref{sec:pair}), pruning them should not hurt task scores---and we in fact observe scattered cases where scores rise.
We verify this on worst-case \textbf{retrieval} (needle-in-a-haystack, Retr.KV), \textbf{long-context reference} (MRCR, LongCodeQA, $\infty$Bench, RULER), and \textbf{mathematical reasoning} (GSM8K, chain-of-thought) (Table~\ref{tab:absbench}).

\textbf{Setup.} Conventions of Table~\ref{tab:absbench}: RULER uses its 0--100 official metric, shown as the mean of 4 tasks with the multi-key NIAH broken out; the 5-task means are 35.8 / 35.7. The $\infty$Bench columns give the equal-weight mean of 10 tasks with Math.Find and Retr.KV broken out, and the 12-task means are full 0.418 / ours 0.375 / MInf 0.420. LongCode is the QA subset of LongCodeBench, exhaustive over the buckets that fit the native context, and these benches score paired window-ON/OFF runs of the same samples.

MRCR, LongCodeQA, and the $\infty$Bench columns are scored with the window injected into a vLLM (v0.22.1) attention backend of Qwen2.5-7B-Instruct-1M (greedy, official scoring); RULER and GSM8K use the eager-attention injection of Section~\ref{sec:eval-quality}.

MRCR and the 4 $\infty$Bench tasks PassKey, Number, Retr.KV, and Math.Find are scored with fp32 attention scores, since we identified and fixed a bf16 score-rounding issue that flipped greedy selections on low-margin digit and hex retrieval; the remaining 8 tasks and LongCodeQA are confirmed invariant to the numeric type by a symmetric paired probe with the pre-fixed criterion $|\Delta|>2\,\mathrm{SE}$ and stay bf16.

The QK-pruned row gives the window's input-independent term reduction at each column's median context, example-weighted for $\infty$Bench and over the median input$+$output length for GSM8K. The GSM8K entry divides by the stock implementation's actual term count, since gpt-oss-20b is a hybrid of sliding-window and full-attention layers.

MInference (MInf) values are measured on the official Qwen2.5-1M inference stack; within the native regime DCA reduces to identity, so they are effectively MInference alone, and because the serving stack differs, absolute levels are not always directly comparable to full/ours. The ours$+$MInf row overlays the window on MInference's retained cells and prunes a further 43--50\% of the retained terms.

\textbf{Results.} On LongBench-v2 4-choice QA (Table~\ref{tab:longbench}), the windowed accuracy exceeds full attention in all 5 Qwen2.5-0.5B bands, including the 2 extrapolation bands, while Llama-3.2-3B is higher in 1 of its 4 bands, equal in 2, and lower in 1; given the small per-band samples we read this as a tendency, not a significant margin.
Across all 6 non-extrapolation conditions the same-answer rate is 0.88--0.96: decisions, not just distributions, are preserved.

The worst case for windowing is retrieval of a specific fact placed far away.
Embedding a needle at 5 depths in real LongBench-v2 text and measuring recall with greedy generation, at both $N{=}8$K and 16K the per-wavelength window ($k{=}1$--$8$) achieves \textbf{100\% recall at all policies and all depths} (compute ratio 0.57--0.76), whereas sliding[256] (compute ratio 0.03) is \textbf{0\% at all depths}.
This gap is not explained by the amount of reduction: even in the iso-compute control with the compute ratio matched to 0.45, the sliding window's KL is \textbf{more than two orders of magnitude} larger; within the same budget it reaches only the most recent 44\% of the context, whereas the per-wavelength window covers all distances.
On the official RULER metric the window stays level with full attention; only point-retrieval MK-NIAH (a standard test distinct from the needle experiment above) drops at $k{=}2$ and recovers under the variable-$k$ profile (Table~\ref{tab:absbench}).
Applying the window only to gpt-oss-20b's full-attention layers, GSM8K accuracy stays on par with the unwindowed model through multi-step reasoning (Table~\ref{tab:absbench}; answer match 95.2\%).

\textbf{Benchmark scoring at the 7B model scale.}
We additionally score the window on OpenAI-MRCR, a multi-needle co-reference retrieval benchmark, and on LongCodeQA, a code-domain 4-choice benchmark; where the window is inactive---16K---the ON/OFF scores coincide exactly.
Pooled over all bands (native $n{=}165$ pairs; Table~\ref{tab:absbench}), the window broadly preserves the score (full 0.204 / ours 0.181, not significant). These 165 rows are drawn deterministically and score-blind from all 583 native-length rows, and all four arms score the identical set.
Paired tests (Wilcoxon signed-rank for continuous scores, exact McNemar for accuracies) find a significant drop for the window on only 1 task, Retr.KV ($p{=}0.004$); MRCR ($p{=}0.98$), LongCodeQA, Math.Find ($p{=}1.0$), and all remaining paired tasks show no significant difference, consistent with score preservation. This task is a point lookup over sequences with no semantic structure---the boundary case for the hypothesis of Section~\ref{sec:pair}, which presumes positional information contributes. At the $k{=}2$ default this costs accuracy; the ours 0.30 versus full 0.75 is not a rounding artifact but the window pruning retrieved keys. The loss is a compute setting rather than a hard limit: at uniform $k{=}64$, still pruning 29.0\% of the terms, Retr.KV returns to the full-attention score ($0.300{\to}0.750$; McNemar $p{=}0.004$), while the variable profile, which keeps low-frequency pairs near $k{=}2$, does not recover it (0.350)---what matters is long-range reach, not high-frequency precision.
\section{Evaluation of Speed}
\label{sec:speed}

\textbf{Setup.} The main measurement platform is the RTX PRO 6000 (Blackwell generation sm\_120, 96\,GB GDDR7)---a representative single-GPU inference platform outside the specialization targets of FA3/FA4.
Measurement is a single request (batch 1); timing is the median of CUDA events, and decode is set up to eliminate the apparent speedup of L2 residency.
The only FA-class implementations available as public wheels for this machine are the cuDNN backend of SDPA (PyTorch's scaled dot-product attention) and FlashInfer (the public FA2--FA4 wheels do not support sm\_120, but FA4 runs when built from source, and this build---at the same commit as the port base in Section~\ref{sec:impl}---serves as stock FA4), and we compare against the per-case $\min$(cuDNN, FlashInfer), the \textbf{fastest available FA-class implementation}.
Each port is measured as latency normalized to its stock host; because the window-off path is bit-identical to stock, this ratio isolates the effect of the window itself.

\textbf{Window-table basis.}
The quality experiments (Sections~\ref{sec:eval-quality}--\ref{sec:eval-ability}) use the effective wavelength $\lambda'_r$, whereas the speed measurements use the raw RoPE base. The two coincide for Qwen2.5-7B-1M, which has no rope scaling; for Llama, which has llama3-style scaling, this amounts to stronger pruning than $k{=}2$ on the effective basis; the resulting end-to-end quality and its recovery by one step of $k$ are reported in Section~\ref{sec:speed-e2e}.%

\subsection{Measured Performance of the Ports}
\label{sec:eval-speed}

\begin{figure}[t]
\centering
\includegraphics[width=0.9\columnwidth]{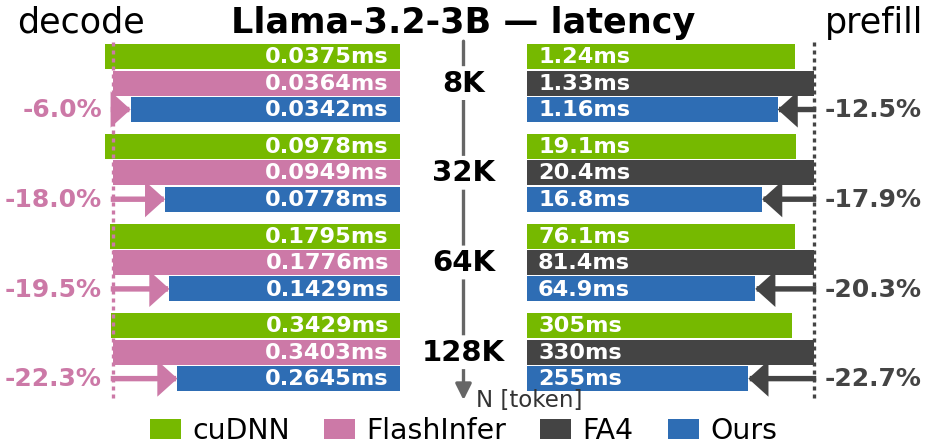}%
\caption{Latency of the two ports (Llama-3.2-3B, same run, lower is better). Left: decode; right: prefill. Bars are relative to each stock implementation (FlashInfer/FA4); dashed lines mark stock ($=1.0$). Values inside the bars are measured times; arrows give the reduction vs stock.}
\label{fig:ports}
\end{figure}

\textbf{Decode: the FlashInfer port.}
With the window on, FlashInfer decode surpasses stock FlashInfer at all evaluated Llama context lengths (latency reductions of 6.0--22.3\%; Figure~\ref{fig:ports}, left).
At a 160K shape outside the native context, decode keeps growing ($1.296\times$ over stock) while prefill holds at the 128K level ($1.193\times$ over the per-case best).
On Qwen2.5-0.5B (2 KV heads, $d{=}64$) the K read does not dominate decode bandwidth and the window has no effect ($1.00\times$).%

\textbf{Prefill: the FA4 port.}
With the window on, FA4 surpasses stock FA4 at all evaluated context lengths, and the reduction (the arrows in Figure~\ref{fig:ports}, right) grows with $N$ (Llama 12.5$\to$22.7\%).
On Llama, moreover, both ports run below cuDNN at every context length (prefill by 6.3--16.4\%).
On the Qwen2.5-7B-1M shape, the same shape as the 1M run, the picture is sharper still: the windowed kernel is the fastest of the 4 systems at every $N$ including 8K, with the reduction growing from 10.4\% at 8K to 29.9\% at 512K.
At 512K the gain over stock ($1.43\times$) exceeds the theoretical ceiling $2/(1+s){=}1.35$ ($s$ $=$ the QK slice ratio of the implemented window table) because the 3 full-attention systems degrade super-quadratically there while the window stays quadratic; the window's intrinsic recovery is 0.77--0.83 at 8K--256K.

\textbf{Agreement of recovery rates.}
Recovery of the theoretical ceiling lines up across the 2 implementation families (CUDA templates, CuTeDSL) on Llama---85--91\% (FlashInfer port) and 92--96\% (FA4 port)---showing that the same methodology of minimal insertion, bit-identical window-off, and prefix-limited loads converts the bulk of the theoretical reduction into wall-clock time.

\subsection{End-to-End Whole-Request Time (Llama)}%
\label{sec:speed-e2e}
\textbf{Setup.} We integrate the two ports into a single inference pipeline, where the KV cache written by the FA4-port prefill is read by the FlashInfer-port decode with no conversion, and measure whole-request time with the window switched on and off in the same binary (LongBench-v2 long documents truncated to each KV size, $n{=}8$).

\textbf{Results.} The saving grows monotonically with context length (Table~\ref{tab:kdial}).
On this extreme-long-document pool the ON/OFF top-1 match falls from 0.90--0.95 at 8K--64K to 0.76 at 128K, but one notch of the retained-period count $k$ recovers the quality.
The cost is about 1\% of whole-request time, demonstrating end to end that $k$ acts as the deployment-time dial between accuracy and compute (Section~\ref{sec:window}).

\begin{table}[t]
\centering
{\footnotesize%
\setlength{\tabcolsep}{1.5pt}
\begin{tabular}{lllll}
\toprule
context & \multicolumn{2}{c}{64K} & \multicolumn{2}{c}{128K} \\%
\cmidrule(lr){2-3}\cmidrule(lr){4-5}
$k$ & \multicolumn{1}{c}{2} & prefill$\to$4 & \multicolumn{1}{c}{2} & decode$\to$3 \\
\midrule
top-1 & 0.92 & \textbf{0.97} & 0.76 & \textbf{0.84} \\
KL & 0.037 & \textbf{0.016} & 0.281 & \textbf{0.148} \\
e2e (prefill/decode) & 1.11 & 1.11 & 1.17 (1.19/1.06) & 1.16 \\
\bottomrule
\end{tabular}}
\caption{The $k$ dial (Llama-3.2-3B, LongBench-v2, $n{=}8$): where quality drops, one notch of $k$ on the named side recovers it at about 1\% of whole-request time; ratios are OFF/ON.}%
\label{tab:kdial}
\end{table}

\begin{table}[t]
\centering
{\footnotesize%
\setlength{\tabcolsep}{3pt}
\begin{tabular}{lllll}
\toprule
$N$ & ceiling $2/1{+}s$ & prefill & decode & whole-request (time) \\%
\midrule
256K & $\sim$1.31$\times$ & 1.18$\times$ & 1.05$\times$ & 1.17$\times$ (58.8$\to$50.2\,s) \\
512K & $\sim$1.35$\times$ & 1.22$\times$ & 1.05$\times$ & 1.22$\times$ (207.5$\to$168.9\,s) \\%
1M & $\sim$1.40$\times$ & \textbf{1.31$\times$} & 1.06$\times$ & \textbf{1.31$\times$} (766.8$\to$587.2\,s) \\
\bottomrule
\end{tabular}}
\caption{End-to-end whole-request time on Qwen2.5-7B-Instruct-1M (window ON/OFF in the same binary; 64K-chunked prefill; $n{=}6$ at every $N$). Ratios are OFF/ON; parentheses give measured seconds (OFF$\to$ON). The maximum per-request deviation from the median is ${\le}\pm2.5$\,s; run-to-run deviation is ${\le}0.23\%$ at 256K/512K (3 runs).}
\label{tab:c2e2e}
\end{table}

\subsection{Scaling to 1M Contexts (Qwen2.5-7B-1M)}
\label{sec:speed-1m}
\textbf{Setup.} To probe the regime where the logarithmic reduction is deepest, we run the same pipeline on Qwen2.5-7B-Instruct-1M (native context 262{,}144, RoPE base $10^7$, no rope scaling), with the window tables regenerated from its effective frequencies and the prefill executed in 64K chunks (bit-identical to single-shot prefill, both arms).

\textbf{Results.} On LongBench-v2 long documents truncated to 256K/512K/1M tokens, the ON/OFF gains grow monotonically with context, reaching a $1.31\times$ whole-request speedup at 1M (Table~\ref{tab:c2e2e}).
The gap between the attention-only theoretical ceiling $2/(1+s)$ and the measured prefill gains reflects non-attention compute and narrows at longer contexts as attention dominates.
Quality holds at 256K (native) and 512K: top-1 is 0.996/0.992, and the LongBench-v2 answer accuracy at 256K is identical ON and OFF.
1M lies $\times$3.85 beyond the native context; top-1 falls to 0.77 there, which we report as extrapolation behavior.

\section{Conclusion}
\label{sec:conclusion}

\textbf{Limitations.}
(i)~The window is fixed in advance, and input adaptation is out of scope.
(ii)~The theoretical reduction rate is a term-count proxy (Section~\ref{sec:theory}); wall-clock time is complemented by measurement.
(iii)~The speed measurements are on a representative platform of a single hardware generation (sm\_120, GDDR7); generalization to other generations is unverified.
(iv)~Quality verification centers on teacher-forced metrics. At the default $k{=}2$ one $\infty$Bench task degrades significantly (Retr.KV), retrieval over structureless strings, and uniform $k{=}64$---still pruning 29\% of the terms---recovers it (Section~\ref{sec:eval-ability}), so $k$ trades computation against this worst case rather than fixing a hard limit.
(v)~Kernel speed is median-based without error bars (end-to-end deviations: Table~\ref{tab:c2e2e}).
(vi)~We do not compare directly with the frequency-exploiting sparse methods or the sink-augmented window~\citep{streamingllm}.

\textbf{Future work.}
Main directions are $V$-side reduction via block-sparsification and content-addressed management of unrotated KV toward multi-request inference.

\textbf{Conclusion.}
The ports into the released FA4 and FlashInfer surpass their stock implementations at all evaluated Llama context lengths, and end to end the whole-request speedup reaches $1.31\times$ at a 1M-token context (Table~\ref{tab:c2e2e}).

\section*{Acknowledgments}
This work was supported by the Joint Usage/Research Center for
Interdisciplinary Large-scale Information Infrastructures (JHPCN) and the
High Performance Computing Infrastructure (HPCI) in Japan, under project ID
jh260017. This work was also supported by JST, the Next Generation Edge AI
Semiconductor Research and Development Project, Grant Number JPMJES2511.
The experiments in this paper used the supercomputer Miyabi (OFP-II) at the
Joint Center for Advanced High Performance Computing (JCAHPC), the
supercomputer Genkai at the Research Institute for Information Technology,
Kyushu University, and the computational resources of R-CCS Cloud provided by
the RIKEN Center for Computational Science (R-CCS).

\bibliography{atflash_arxiv_v1}

\clearpage
\appendix
\twocolumn[{%
  \begin{center}
  {\LARGE\bfseries Appendix\par}
  \vspace{0.8em}
  \end{center}
}]
\noindent
This appendix records (A) a family of window rules of which the
per-RoPE-wavelength distance window is one point, (B) the distinction between
the cache footprint, which position-independent caching shrinks, and the
read-time rotation count, which the window prunes, (C) the closed form of the
reduction rate, including the decode case that the main text states without
writing out, (D) the two implementation routes the pair decomposition admits,
and why we take the one that keeps an FA-class kernel intact, and (E) notes on measurement and verification.

\section{Generalizing the window rule}
We use the notation of the main text throughout ($d$, the pair index $r$, the
wavelength $\lambda_r$, the RoPE base $B$, the context length $N$, the per-pair
window width $w_r$).

\begin{figure}[t]
\centering
\includegraphics[width=\columnwidth]{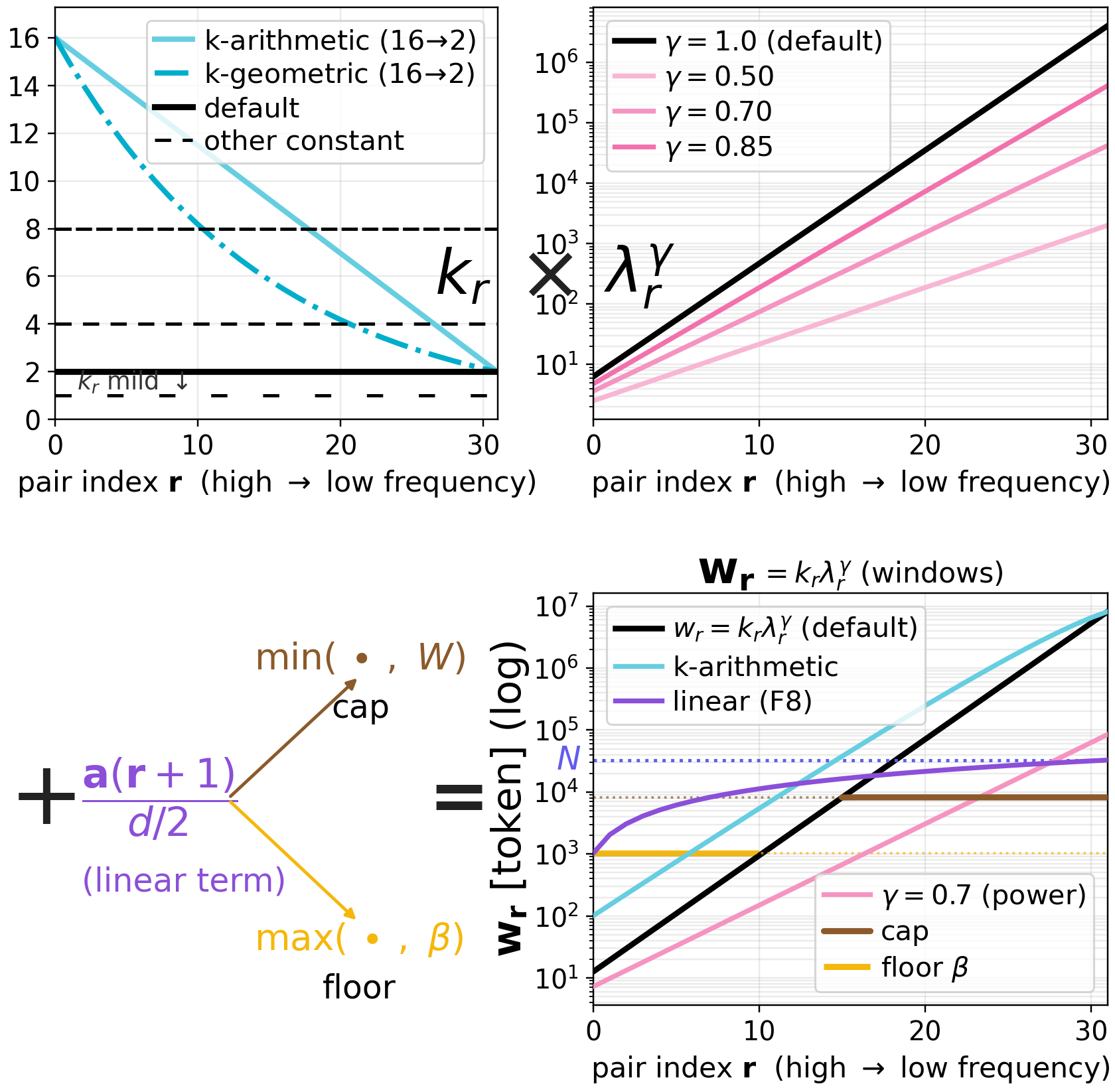}
\caption{Factorization of the window pattern into a retention profile, a power of
the RoPE wavelength, and floor/ceiling terms.}
\label{fig:family}
\end{figure}

The default window $w_r=\min(k\lambda_r,N)$ of the main text is one point of a
larger family. Factorizing a window into a retention profile $k_r$ times a power
of the wavelength, clamped between a floor and a ceiling, gives
\begin{equation}
w_r=\min\!\bigl(\max(k_r\,\lambda_r^{\gamma}+a\,\tfrac{r+1}{d/2},\ \beta),\ W\bigr)
\label{eq:family}
\end{equation}
where $k_r$ is the retention profile (a constant $k$, or a gradient that keeps
more periods at high frequencies), $\gamma$ is the wavelength exponent, $\beta$
is a floor in tokens that keeps near distances in full, $W$ is a ceiling, and
$a\,\tfrac{r+1}{d/2}$ is an additive term that is \emph{linear in the pair index
$r$} and carries no wavelength dependence at all; it is $0$ in most settings. The default window is the special case
$k_r{=}k,\ \gamma{=}1,\ \beta{=}0,\ W{=}N,\ a{=}0$; setting $k_r{=}0,\ a{=}W$
degenerates to a \emph{linear} window, which the figures label as such.
Figure~\ref{fig:family} illustrates the factorization panel by panel.

The retention profile is either constant or interpolates from a high-frequency
value $k_{\mathrm{hi}}$ down to a low-frequency value $k_{\mathrm{lo}}$. We sweep
two interpolations, and the figures name them accordingly:
\emph{$k$-arithmetic} decreases $k_r$ by a constant amount per pair, linear in
$r$; the variable-$k$ profile of the main text is this case with
$k_r=16-\tfrac{14}{d/2-1}\,r$, running from $16$ at the highest frequency to $2$
at the lowest. \emph{$k$-geometric} instead decreases $k_r$ by a constant ratio
per pair, so it falls faster at high frequencies. The other labels follow the
same equation: $cN$ is a ceiling set as a fraction of the sequence length, and
$n\%$ is the fraction of the lowest pairs made position-free. \emph{Sliding $W$}
is the sliding window, the special case in which every pair receives the same
width.

\paragraph{Position-free pairs as a limit.}
A pair whose rotation is removed (NoPE) has no wavelength, so no distance
renders it positionally uninformative: it is the limit $\theta_r\!\to\!0$, in
which $w_r\ge N$ for every $N$ and the pair is retained permanently. The fraction of
low-frequency pairs treated this way is a separate knob from the window
factorization itself. Architectures that interleave rotated and unrotated
attention layers therefore fall inside the same description: the rotated layers
are pruned by Eq.~(\ref{eq:family}) and the unrotated ones are not pruned at all.

\begin{figure*}[t]
\centering
\includegraphics[width=\textwidth]{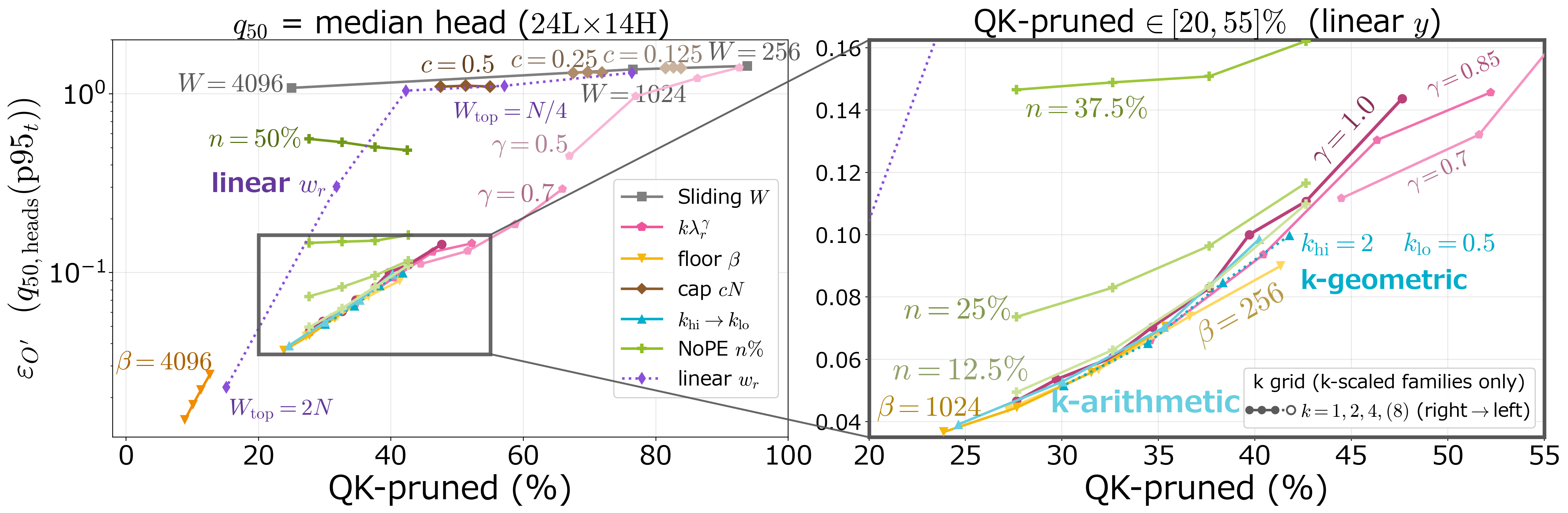}
\caption{Error of the window families of Eq.~(\ref{eq:family}) against the
fraction of inner-product terms pruned (QK-pruned); Qwen2.5-0.5B, $N{=}8192$.
Left: the whole sweep on a logarithmic error axis; right: the practical range on
a linear axis. Series and marker conventions are given by the in-figure legends;
the families are defined in the text.}
\label{fig:families}
\end{figure*}

\paragraph{Sweeping the family.}
Figure~\ref{fig:families} sweeps the knobs of Eq.~(\ref{eq:family}) against the
compute they save, measured as the fraction of inner-product terms pruned, the
quantity the main text reports as QK-pruned. The sweep runs on
Qwen2.5-0.5B at $N{=}8192$; the error axis is the relative
deviation of the windowed attention output from full attention, taken per head
as the 95th percentile over query positions and summarized by the median head.
This deviation is measured on the attention output itself; closeness of the
model's output distribution and task scores are established separately in the
main text.
Three things are visible. The sliding window is dominated everywhere: at any
level of pruning it carries an error far above every wavelength-scaled setting.
The wavelength-scaled families---the exponent $\gamma$, the floor $\beta$, the
cap $cN$, and a gradient in $k_r$---collapse onto a common front over the
practical range, so within this setting the additional knobs buy little over the plain
$\gamma{=}1$ default that the main text uses. Making a fraction of the
low-frequency pairs position-free costs error at equal compute, which is the
expected price of removing the very components that carry long-range position.

\begin{figure}[t]
\centering
\includegraphics[width=\columnwidth]{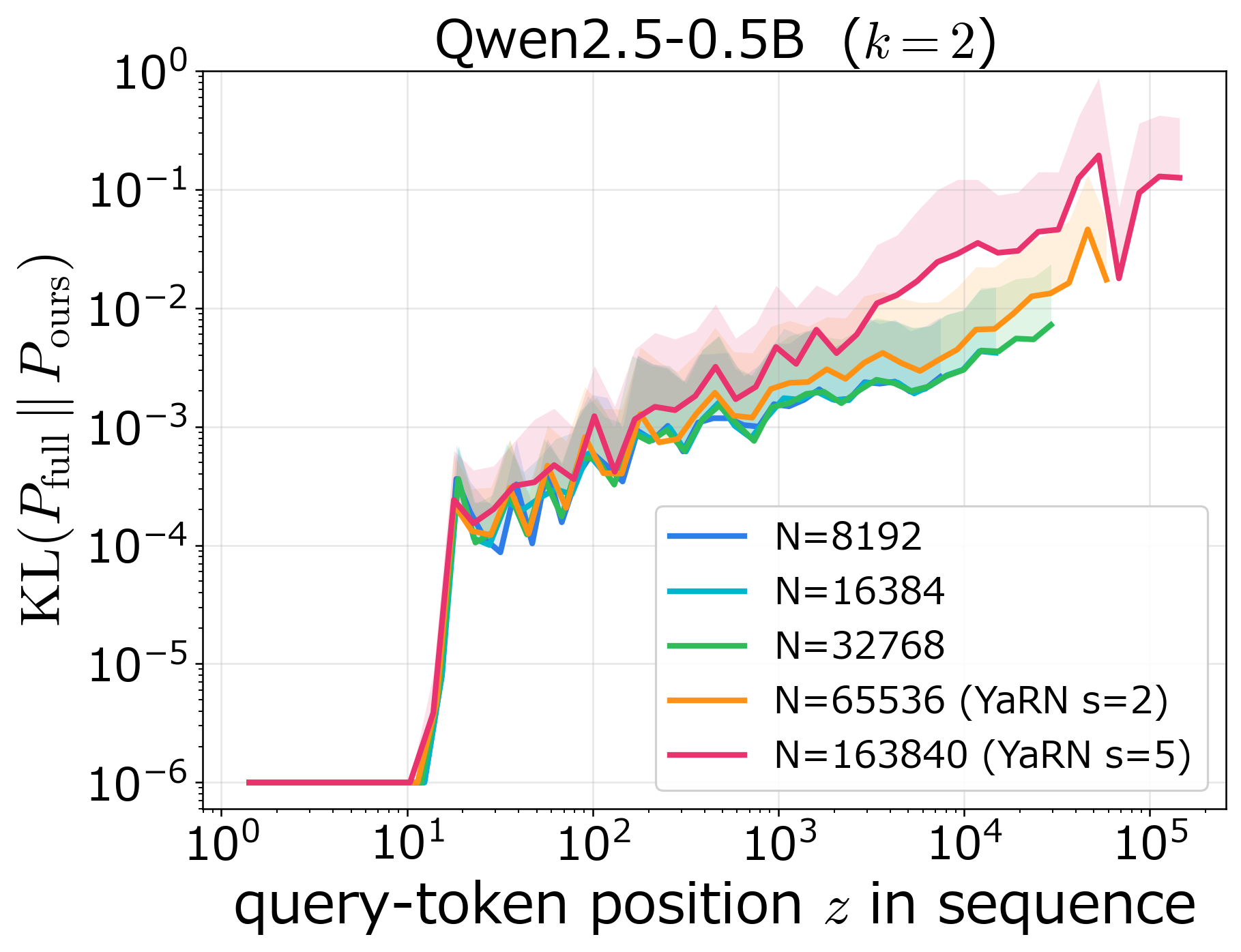}
\caption{KL divergence from full attention against query position in the
sequence, per context length. Solid lines give the mean per-token KL within each
position bin; the shaded band above each line extends from that mean to the 95th
percentile of the same bin (the upper tail).}
\label{fig:posprof}
\end{figure}

\paragraph{Fidelity along the sequence.}
The main text reports the mean KL divergence per band. Figure~\ref{fig:posprof}
resolves it along the query position: within the non-extrapolation regime the
curves stay in the $10^{-3}$\,nat range, and only the two YaRN-extended settings
depart from it.

\section{Cache footprint versus rotation count}
Position-independent caching (MEPIC~\citep{mepic}) keeps the
KV cache \emph{unrotated} and fuses the rotation into attention at read time;
this shrinks the memory the cache must hold---its \emph{footprint}---because
entries carry no position yet and can be shared or deduplicated across requests.
The per-wavelength window does not shrink the cache. What it reduces is the
number of rotations that have to be applied at read time: out-of-window pair
components are never brought into on-chip memory (SRAM), so the corresponding
rotations---and the trigonometric evaluations or table lookups behind them---are
not performed either.

The two reductions therefore act on different axes of the same runtime-fused
path and can be applied together. The main text measures the analogous
composition with token-level sparsity: overlaying the window on the cells kept by
a dynamic-sparse method prunes a further 43--50\% of the retained terms.

In the single-request, single-GPU setting of our speed measurements the
trigonometric work is not the bottleneck, which is why the main text quantifies
compute in inner-product terms and in transferred bytes. Multi-request serving
by itself does not change this either: per request, the rotation work still
grows only linearly. What changes the picture is the position-independent cache
itself. There the cache holds \emph{one unrotated entry per token}, shared
across the requests that reference it, and the rotation is fused into every
read: the trigonometric work then scales not with the tokens written but with
the reads of that entry, summed over all sharing
requests~\citep{mepic}. Reads dwarf writes in exactly the settings that motivate
such caches, and it is this read-time rotation that the per-wavelength window
prunes pair by pair. That regime is outside the scope of this paper.

\section{Deriving the reduction rate}
The main text solves the double integral over rows and pair index for the whole
prefill. It is worth solving it once for an arbitrary row interval, because
prefill and decode then become the same expression evaluated on different rows
rather than two separate derivations. We then carry the same integral through the
window family of Eq.~(\ref{eq:family}).

\begin{figure}[t]
\centering
\includegraphics[width=0.98\columnwidth]{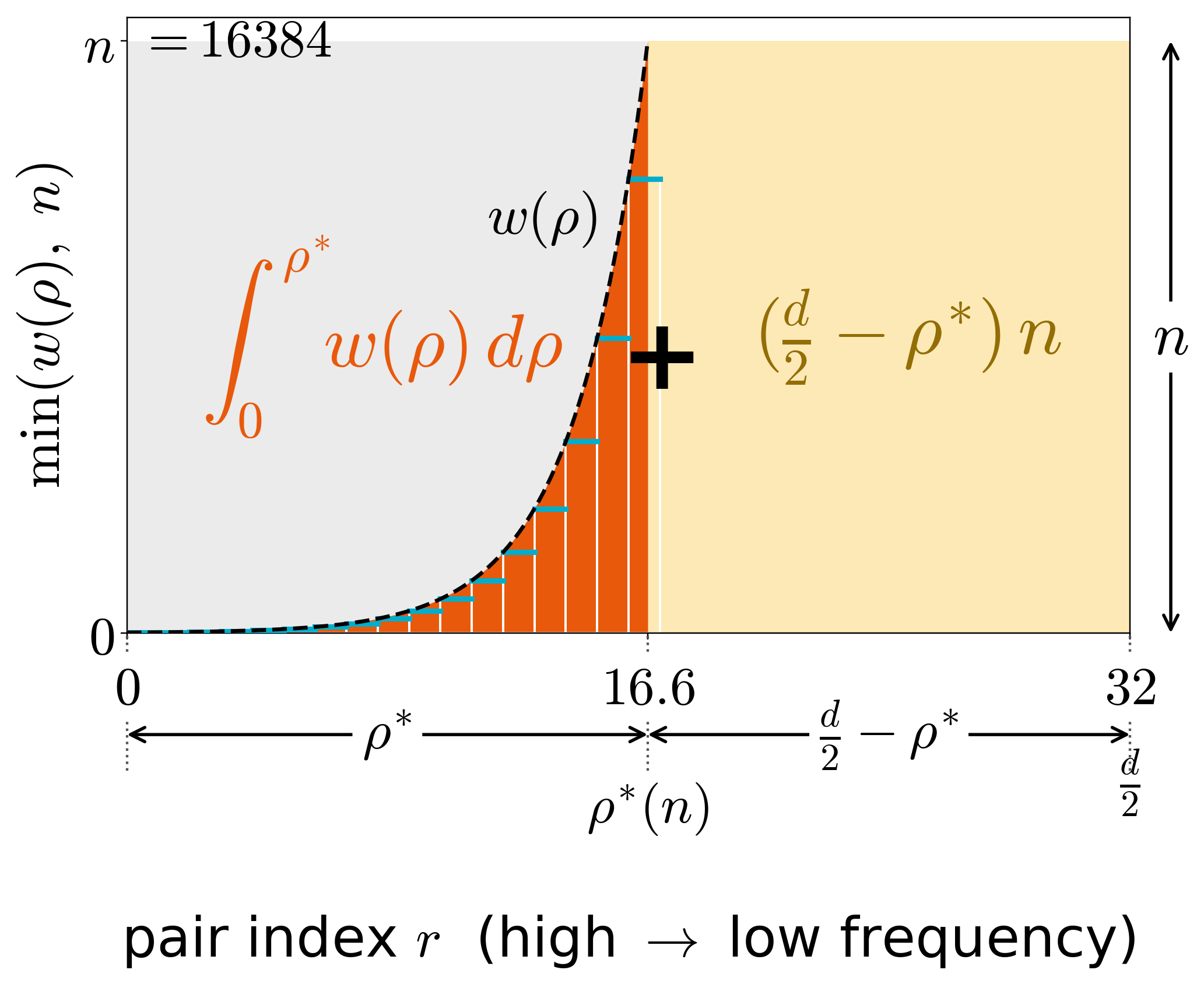}
\caption{The split of Eq.~(\ref{eq:Kn}) for a single query row, to scale
(Qwen2.5-0.5B constants, $k{=}2$, $n{=}16384$). The painted areas equal the
two terms: orange under the dashed continuous $w(\rho)$ up to
$\rho^{*}(n){=}16.6$, yellow beyond; gray: pruned. The cyan steps mark the
discrete $\min(w_r,n)$ actually counted; their deviation from the curve is the
integral-approximation error (pruned: 46.3\% discrete, 44.7\% continuous).}
\label{fig:sndecomp}
\end{figure}

\paragraph{Step 1: an arbitrary row interval.}
Following the main text we relax the discrete pair index $r$ to a continuous
variable $\rho$; for the default window ($k_r{=}k$, $\gamma{=}1$) this gives
$w(\rho)=w_{\min}B^{2\rho/d}$ with $w_{\min}=2\pi k$. Let $s(n)$ denote the
number of inner-product terms the window retains in query row $n$, the integrand
of the main text's double integral $S$ (the letter $K$ is kept for the
truncation order of Appendix~\ref{sec:tworoutes}):
\begin{equation}
s(n)=\int_{0}^{d/2}\min\bigl(w(\rho),n\bigr)\,d\rho .
\label{eq:sdef}
\end{equation}
Full attention computes $\tfrac{d}{2}n$ terms in the same row.
For $n\ge w_{\min}$ the minimum switches at the saturation boundary
$\rho^{*}(n)=\tfrac{d}{2}\ln(n/w_{\min})/\ln B$, where $w(\rho^{*}){=}n$, so
\begin{equation}
s(n)=\underbrace{\biggl(\int_{0}^{\rho^{*}}\!\!w(\rho)\,d\rho\biggr)}_{\rho<\rho^{*}(n)}
\;+\;\underbrace{\Bigl(\frac{d}{2}-\rho^{*}(n)\Bigr)n}_{\rho\ge\rho^{*}(n)} .
\label{eq:Kn}
\end{equation}
The first term integrates the window widths of the pairs left of the boundary
and evaluates to $\tfrac{d}{2\ln B}(n-w_{\min})$; the second counts the
remaining pairs, for which the causal length $n$ binds instead.
Figure~\ref{fig:sndecomp} draws the split to scale for a single query row, each
area labeled by its own term.

Integrating the pruned count $\tfrac{d}{2}n-s(n)$ over the rows $[n_0,n_1]$ and
dividing by the $\tfrac{d}{2}\cdot\tfrac{n_1^{2}-n_0^{2}}{2}$ terms that full
attention holds there gives the reduction rate of that interval, which we denote
$R(n_0,n_1)$. Writing
\begin{equation}
F(n)=n^{2}\Bigl(\ln\frac{n}{w_{\min}}-\frac32\Bigr)+2w_{\min}n
\label{eq:Fdef}
\end{equation}
for the primitive that this integration produces (up to constant factors),
\begin{equation}
R(n_0,n_1)=\frac{\bigl[\,F(n)\,\bigr]_{n_0}^{n_1}}{(n_1^{2}-n_0^{2})\,\ln B}\,.
\label{eq:Rgen}
\end{equation}

\paragraph{Step 2: prefill and decode as substitutions.}
Both cases of the main text are substitutions into Eq.~(\ref{eq:Rgen}), and each
constant can be read straight off $F$. The whole prefill is
$(n_0,n_1)=(1,N)$, where $F(1)$ is negligible and $F(N)/(N^{2}\ln B)$ remains; dropping the
$2w_{\min}N$ term for $N\gg w_{\min}$ leaves the prefill form with exactly the
constant the primitive carries, $-\tfrac32$. Decode is the last row alone,
$(n_0,n_1)=(N{-}1,N)$: the bracket is then the unit-step difference
$F(N)-F(N{-}1)$, which is $F'(N)$ to leading order, with
\begin{equation}
F'(n)=2n\bigl(\ln\tfrac{n}{w_{\min}}-1\bigr)+2w_{\min} ,
\label{eq:Fprime}
\end{equation}
and division by the denominator $(2N{-}1)\ln B\simeq 2N\ln B$ leaves $-1$. In
the notation of the main text, as $N\to\infty$ below saturation,
\begin{equation}
\begin{aligned}
\mathrm{reduction}_{\mathrm{prefill}}&=R(1,N)\\
&=\frac{\ln(N/w_{\min})-3/2}{\ln B}+O\bigl(w_{\min}/N\bigr),\\
\mathrm{reduction}_{\mathrm{decode}}&=R(N{-}1,N)\\
&=\frac{\ln(N/w_{\min})-1}{\ln B}+O\bigl(w_{\min}/N\bigr).
\end{aligned}
\label{eq:reddecode-apx}
\end{equation}
The two constants are therefore not separate results: the whole prefill reads the
primitive, decode reads its derivative. The $O(w_{\min}/N)$ terms can be written
out:
evaluating the last row directly, with no outer integration, gives
$1-s(N)/(\tfrac{d}{2}N)=\bigl[\ln(N/w_{\min})-1+w_{\min}/N\bigr]/\ln B$, so the
decode error is exactly $w_{\min}/(N\ln B)$, and the prefill form carries
$2w_{\min}/N$ in the same place. Neither leading term is a limit value: the
rates rise along the logarithmic law until saturation ends it, and the only value
the reduction converges to as $N\to\infty$ is $1$ (Step 3).

\paragraph{The interval itself, not only its two extremes.}
Keeping the general interval has a direct use. In a
long-running session, a further user message or a returned tool result is
appended on top of a context that is already long, and with the existing KV
retained, only the new rows are computed. Each such step
evaluates $[n_0,n_1]$ with $n_0$ the length already present, which is
Eq.~(\ref{eq:Rgen}) with neither extreme substituted---the whole prefill
($n_0{=}1$) and decode ($n_1{-}n_0{=}1$) are its endpoints. The reduction rises as
$n_0$ moves up the context, because the short early rows, the ones no window can
prune, are no longer counted: over the second half of the context, $[N/2,N]$,
Eq.~(\ref{eq:Rgen}) gives, to the same order,
$\bigl(\ln(N/w_{\min})-1.27\bigr)/\ln B$, about 1.7
points above the whole prefill at $B{=}10^{6}$, or half of the 3.6-point gap to
decode. (This is distinct from chunked prefill, which splits one input but still
computes every row from 1 to $N$, so its total remains $R(1,N)$.)

\paragraph{Step 3: what converges as $N$ grows.}
Two things do. First, both intervals share the slope $1/\ln B$, so their
\emph{gap} converges to the constant $(\tfrac32-1)/\ln B=1/(2\ln B)$---the 3.6
points quoted in the main text at $B{=}10^{6}$. Decode is the deeper of the two
because the prefill value averages in the early rows, which are still short
enough for the window to cover them. Second, the logarithmic law itself holds only
until the boundary saturates: at $N=k\lambda_{\max}=w_{\min}B$ we have
$\rho^{*}{=}d/2$, no window is clipped by $n$, and Eq.~(\ref{eq:Kn}) collapses to
$s=\sum_r w_r$, independent of $n$. The decode reduction reaches $1-1/\ln B$ there
(92.8\% at $B{=}10^{6}$) and tends to $1$ beyond it, since the retained count stays
fixed while the row keeps growing.

\paragraph{Step 4: the same integral over the general family.}
Carried through the same steps, Eq.~(\ref{eq:family}) shows that the wavelength
exponent reduces to the same law. With
$w(\rho)=k\lambda(\rho)^{\gamma}$ and $\lambda(\rho)=2\pi B^{2\rho/d}$, the window
is again a geometric progression, $w(\rho)=k(2\pi)^{\gamma}B^{2\gamma\rho/d}$, so
every expression above survives under the two replacements
\begin{equation}
\ln B\ \longrightarrow\ \gamma\ln B ,\qquad
w_{\min}\ \longrightarrow\ k(2\pi)^{\gamma} .
\label{eq:gammasub}
\end{equation}
The exponent therefore acts exactly as a rescaling of the RoPE base to
$B^{\gamma}$: a $\gamma$-window on a model of base $B$ prunes like the default
window on a model of base $B^{\gamma}$. Saturation moves to
$N=k\lambda_{\max}^{\gamma}$, where the decode reduction is $1-1/(\gamma\ln B)$.

The other end of the family behaves differently in kind. The degenerate linear
window ($k_r{=}0$, $a{=}W$, i.e.\ $w(\rho)=W(\rho{+}1)/(d/2)$) saturates at
$\rho^{*}=\tfrac{d}{2}(n/W)-1$, and the same two integrals give a retention rate
$1-n/(2W)$ up to $O(1/d)$ corrections. Its reduction is therefore \emph{linear} in
$n$ rather than logarithmic: it reaches one half at $n{=}W$ and then follows
$1-W/(2n)$ once every window is saturated. Both ends tend to full pruning, but the
wavelength-scaled family approaches it logarithmically while the linear window
approaches it as $1/n$---which is the structural reason the two are not
interchangeable at a fixed compute budget.

The floor $\beta$ and the ceiling $W$ keep the integral elementary; they only add
breakpoints where $w(\rho)$ meets them, splitting $[0,d/2]$ into further
subintervals of the same two kinds. The one exception is the mixed case: when
the exponential and the additive linear term are both active, $w(\rho^{*}){=}n$
is transcendental in $\rho^{*}$, and its root is a value of the Lambert function
$W_{0}$ (not the window ceiling $W$). The family is thus elementary at either
end but not in between.

\section{Two routes: slicing and accumulation}
\label{sec:tworoutes}
The exact pair decomposition admits two implementation routes. One slices the
reduction axis and leaves the surrounding kernel alone; the other turns the sum
over pairs into a recurrence and accumulates the exponential along it. We take
only the first, and this section records why.

\paragraph{(a) Slicing the reduction axis (the main text).}
Because the window is static, the set of active pairs per tile is known at
compile time, so the tiling, the running maximum, and the normalizer of an
FA-class kernel are left untouched and only the effective inner-product length
shrinks. Nothing about the model is assumed beyond the RoPE constants.

\begin{figure}[t]
\centering
\includegraphics[width=\columnwidth]{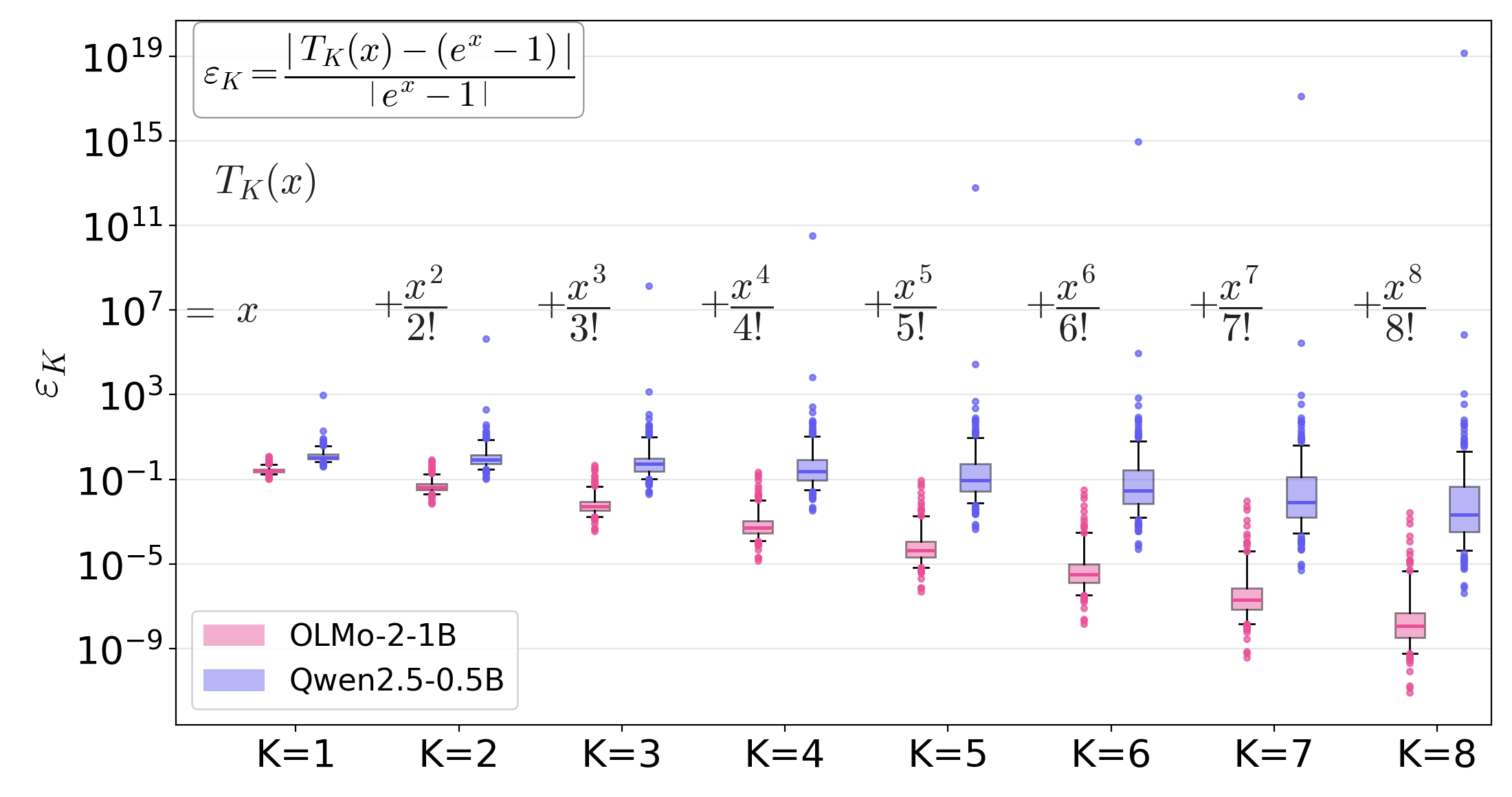}
\caption{Relative error of the truncated increment against the truncation order,
for a model with query--key normalization (OLMo-2-1B) and for one without
(Qwen2.5-0.5B).}
\label{fig:qknorm}
\end{figure}

\paragraph{(b) Accumulating the exponential in increment form.}
With the partial sum of the score over pairs written as
$c_{m,r}=c_{m,r-1}+\Delta x_{m,r}$, where $m$ is the key position and
$\Delta x_{m,r}$ the per-pair contribution of the main text's score
decomposition, and with $\ell_{m,r}=e^{c_{m,r}}$, the exponential law gives the
identity
\begin{equation}
\ell_{m,r}=\ell_{m,r-1}+e^{\,c_{m,r-1}}\bigl(e^{\,\Delta x_{m,r}}-1\bigr),
\label{eq:incr}
\end{equation}
which is exact. Each pair then contributes only through the increment
$e^{\Delta x_{m,r}}-1$, which for small arguments can be truncated,
\begin{equation}
e^{\,\Delta x_{m,r}}-1=\Delta x_{m,r}+\frac{\Delta x_{m,r}^{2}}{2!}
   +\frac{\Delta x_{m,r}^{3}}{3!}+\cdots ,
\label{eq:expm1}
\end{equation}
so that the exponential is never evaluated directly. The catch is the range of
$\Delta x$, and it is not a matter of degree. Figure~\ref{fig:qknorm} contrasts a
model that normalizes queries and keys with one that does not. With
normalization the error falls steadily with the order: the median head reaches
$5.4\times10^{-4}$ at $K{=}4$ and $1.1\times10^{-8}$ at $K{=}8$. Without it the
median error is two to three orders of magnitude larger at equal truncation
order ($2.4\times10^{-1}$ at $K{=}4$). More damaging for a kernel, which must be
correct for every head, the worst heads \emph{diverge} as the order grows: the
largest head error rises from $9.4\times10^{2}$ at $K{=}1$ to $1.5\times10^{19}$
at $K{=}8$, and one head overflows outright, because the expansion is applied to
logits of magnitude $\sim\!9.4\times10^{2}$. Route (b) also gives up the
running-maximum subtraction that makes the standard formulation safe, so it is a
different kernel architecture rather than a modification of one.

We take route (a) because it is model-agnostic: it requires no assumption on the
logit range and it preserves bit-identity with the host kernel when the window is
off. Route (b) remains open, and with it the question that both routes
share---how to avoid holding $e^{x}$ per element at all.

\paragraph{Precision as the other side of the compute--accuracy trade.}
Pruning terms shortens the score sum, which can narrow the margins that decide
the arg-max. We observed this directly on the digit-retrieval task Math.Find:
under bf16 the windowed arm's score was depressed while the full-attention arm
was invariant to the numeric type, and rescoring both arms in fp32 closed the
gap; the paired test in the main text returns $p{=}1.0$. Splitting a low-precision product to recover
accuracy is therefore a natural direction for spending part of what the window
saves; we make no performance claim here.

\section{Notes on measurement and verification}
These notes record protocol details that the main text compresses for space;
none of them changes a reported number.

\paragraph{Answer-format addendum for retrieval-style tasks.}
The four retrieval-style $\infty$Bench tasks (PassKey, Number, Retr.KV,
Code.Run) are scored
under a one-line prompt addendum that fixes the answer format.

\paragraph{Two-site execution of $\infty$Bench.}
Byte-identity of the $\infty$Bench data and tokenizer across the main text's
two GPU platforms was verified.

\paragraph{MRCR native sample.}
The recipe behind the main text's draw of the $n{=}165$ native rows: within
each length bin the rows are sorted by length, and contiguous blocks from a
fixed offset are accumulated wave by wave. In the extrapolation band (491--496K
tokens, $n{=}3$, where the dual-chunk attention (DCA) mechanism of the official
serving stack is active) every method scores near zero.

\paragraph{Choice of the official RULER metric.}
On the broad-reference tasks---variable tracking (VT), common-words extraction
(CWE), and frequent-words extraction (FWE)---strict accuracy is zero even for
full attention; the official metric is the one under which the baseline itself
is measurable.

\paragraph{Implementation checks behind bit-identity.}
Beyond bit-identity with the window off, the ports passed: equality of the
window-on path at $k{=}\infty$ with the window-off path; correctness against a
reference implementation; hash-identical kernel outputs before and after rebasing
the window patch (FA4 port); and, for the FlashInfer port, agreement of output
hashes across four configurations on Llama with the window tables rounded, at
injection, to that port's 16-byte vector-load granularity (a load-path constant,
distinct from the slice-boundary rounding of the main text).

\paragraph{Decode timing without L2 residency.}
The decode platform has a nominal bandwidth of 1{,}792\,GB/s and a 128\,MiB L2;
timed reads cycle through a KV pool of at least 512\,MiB.

\end{document}